\documentclass{article}

\usepackage{arxiv}

\usepackage[utf8]{inputenc} 
\usepackage[T1]{fontenc}    
\usepackage{hyperref}       
\usepackage{graphicx}
\usepackage{setspace}
\usepackage{ifpdf}
\usepackage{graphicx}
\usepackage{amsmath}
\usepackage{amssymb}
\usepackage{mathtools}
\usepackage{longtable}
\usepackage{pdfpages}

\usepackage{algorithm}
\usepackage[noend]{algpseudocode}

\usepackage{etoolbox}
\usepackage{multirow}
\usepackage{epsfig}
\usepackage{times}
\usepackage{array}
\usepackage{bm}
\usepackage[figuresright]{rotating}
\usepackage{graphics}
\usepackage{subcaption}

\usepackage{url}
\usepackage{hyperref}
\hypersetup{
    colorlinks,
    citecolor=black,
    filecolor=black,
    linkcolor=black,
    urlcolor=black
}

\newcommand{\sectionref}[1]{Section~{\ref{#1}}}

\usepackage{tikz}
\usetikzlibrary{shapes,fit,arrows,calc,positioning}

\tikzstyle{block1} = [draw, fill=blue!20, rectangle, 
    minimum height=2em, minimum width=4em]
\tikzstyle{block2} = [draw, fill=green!20, rectangle, 
    minimum height=2em, minimum width=4em]
\tikzstyle{block3} = [draw, fill=red!20, rectangle, 
    minimum height=2em, minimum width=4em]

\title{Human action recognition using local two-stream convolutional neural network features and support vector machines}

\author{
 David Torpey \\
  School of Computer Science and Applied Mathematics\\
  University of the Witwatersrand\\
  Johannesburg, South Africa \\
  \texttt{torpey.david93@gmail.com} \\
   \And
 Turgay Celik \\
  School of Coumputing and Information\\
  University of the Witwatersrand\\
  Johannesburg, South Africa \\
  \texttt{celikturgay@gmail.com} \\
}

\begin{document}
\maketitle
\begin{abstract}
This paper proposes a simple yet effective method for human action recognition in video. The proposed method separately extracts local appearance and motion features using state-of-the-art three-dimensional convolutional neural networks from sampled snippets of a video. These local features are then concatenated to form global representations which are then used to train a linear SVM to perform the action classification using full context of the video, as partial context as used in previous works. The videos undergo two simple proposed preprocessing techniques, \emph{optical flow scaling} and \emph{crop filling}. We perform an extensive evaluation on three common benchmark dataset to empirically show the benefit of the SVM, and the two preprocessing steps.
\end{abstract}


\section{Introduction}
Human action recognition is a very active area of study in the computer vision research community. This is likely because a viable, reliable solution to such a problem would have a vast impact on society, in domains such as healthcare, surveillance, and entertainment. In the surveillance space, detection of anomalous or illegal events, such as shoplifting; robbery; and fighting, would be possible. In entertainment applications, human-computer interaction could reach new levels of effectiveness, since reliable detection of behaviour and engagement would be possible. Lastly, in the healthcare industry, a solution could assist in the rehabilitation of patients \cite{ke1}.

In a video, actions are place, and are captured by cameras of different sensors. Finding a general, reliable, and robust solution to (video) human action recognition is still an open problem. Both motion and appearance information needs to be accounted for when modelling the problem, and thus extracting a reliable set of features that generalise to novel settings is difficult. The features, or representations of actions, should be discriminative enough to tell the difference between temporally-similar, and spatially-similar, actions.

Typically, there are two approaches to human action recognition. The first, which dominated early action recognition research, involves extracting hand-engineered features \cite{laptev2,laptev3,grauman1}, henceforth termed \emph{engineered features}. The most common, high-level approach to extracting engineered features starts with an interest point detector - typically either space-time-interest points \cite{laptev1} (STIPs) or dense sampling. Then, feature descriptors are extracted at each of these locations. Such descriptors include, but are not limited to, histograms of oriented gradients \cite{dalal1} (HOG), histograms of optical flow \cite{chaudry1} (HOF), $N$-jets (\cite{laptev2,laptev3}, and motion boundary histograms \cite{mbh1} (MBH). These features are typically computed locally - in a 2D or 3D window around the interest points. These local features are then encoded and quantised to form global, fixed-length representations for the video. The two most common approaches to do this is the bag-of-words framework (BoW), or Fisher vectors. These then serve as the final action representation, which are commonly fed into an SVM to perform the classification.

The second set of approaches, unsurprisingly, leverage deep learning. Simple inputs, such as RGB video or optical flow videos, are fed into 2D or 3D convolutional neural networks (CNNs) to learn salient spatio-temporal features about the actions, and perform the classification. We term these features \emph{automated features}. Up until recently, very few large, labelled action recognition datasets existed, and thus the benefit of deep learning could not be realised in this domain. As such, the approaches extracting engineered features typically dominated the research landscape. However, with the introduction large-scale action recognition datasets, such as Kinetics and YouTube-8M, deep learning has reaps the rewards. Approaches based on deep learning are now achieving state-of-the-art results on all common benchmarks datasets. The issue with such approach is their massive computational resource demands - some state-of-the-art techniques being trained using upwards of 60 GPUs \cite{i3d}.

This work serves to introduce a novel approach to human action recognition, and demonstrates a number of key benefits to aid recognition performance. Firstly, performing two simple pre-processing techniques can assist the networks with learning better action representations, and using a linear SVM trained on a set of \emph{local}, crop-level action representations improves performance greatly from simply taking a consensus vote of the network's crop-level predictions.

The remainder of the work is structured as follows. \sectionref{sec:Background and Related Work} provides a background of the relevant literature, and \sectionref{sec:Methodology} introduces the proposed approach. \sectionref{sec:Experimental Results} provides a comprehensive evaluation of the proposed method on the two benchmark datasets. Lastly, \sectionref{sec:Conclusion and Future Work} concludes the work and provides potential future work.

\section{Background and Related Work}
\label{sec:Background and Related Work}

Previous research into action recognition can largely be split into two groups - those hand-crafting features, and those learning feature automatically using deep neural networks. We split the literature review in accordance with this.
\subsection{Engineering Features}
\cite{laptev2} detect interest points in a video, at various spatial and temporal scales, using an 3D extension of the Harris operator, known as space-time interest points \cite{laptev1} (STIPs). At each of these locations, fourth-order local jets are computed. A BoW approach is taken, clustering a subset of these jets using $K$-Means. The final video representation is then the resultant histogram of visual word occurrences. An SVM was trained on these histograms. This approach resulted in a best average accuracy of 71.7\% on the KTH dataset. \cite{laptev5} takes a similar approach, starting with STIPs. Then, instead of computing $N$-jets at these locations, HOG and HOF vectors are instead computed in space-time volumes around the detected points. These HOG and HOF vectors are concatenated and normalised for each subvolume within this volume. A BoW framework is employed, with $K=4000$. An SVM with a $\chi^2$ kernel is used for classification. This approach resulted in an average accuracy of 91.8\% on the KTH dataset. This method was state-of-the-art at the time, due to the features being able to capture more of the pertinent motion and appearance information than that of \cite{laptev2}, for example. \cite{laptev3} evaluated various interest point and descriptor combinations on a host of datasets. Similar to the approaches above, video sequences are represented as a bag-of-words. The descriptors are quantised into visual words using K-Means clustering with $V=4000$. These visual words are then used to compute histograms of visual word occurrences, and these serve as the final representations for the video. For classification, an SVM is used with the $\chi^2$ kernel. One key finding is that in realistic video settings, dense sampling outperforms other interest point detection methods, while increasing the number of features by a factor of 15. Another key finding is that the HOG/HOF descriptors seem to, in general, perform well. \cite{corso1} introduce \emph{Action Bank} - an extension of the \emph{Object Bank} \cite{li1} approach to image representation into the video domain. This method is the current state-of-the-art on the KTH dataset. Action bank represents a video as a collected output of many action detectors that each produce a correlation volume. Results obtained were 98.2\% on KTH, 95.0\% on UCF Sports, 57.9\% on UCF50, and 26.9\% on HMDB51. The two main downsides of this approach are its massive computational inefficiency (e.g. a video from UCF50 took anywhere from 0.4 - 34 hours), which is likely the reason the method has not been researched further, as well as the fact it performs comparatively poorly on realistic datasets.

\subsection{Automated Features}
A method that spawned much further research in action recognition was introduced by \cite{tscnn}. This method is based on the intuition that video can naturally be broken down into spatial and temporal components. The spatial component, defined by the individual frames of the videos, carries information about scenes and objects in the video. The temporal component, defined by motion between frames, carries information about the movement in the video (e.g. that of camera motion and the objects). Both of these components, or streams, are implemented as CNNs. The spatial stream CNN operates on individual RGB frames of the video. Thus, this network makes action predictions for each frame. The static appearance of a frame is hypothesised to be a useful cue for action recognition since actions are often strongly associated with particular objects. The temporal stream CNN is fed some form of pre-computed optical flow input. This is done so that the temporal network does not have to estimate the motion itself. Three different types of inputs are considered for the temporal CNN. To fuse the output of both the spatial and temporal networks, two methods are considered: averaging and training a linear SVM on stacked $L_2$-normalised softmax vectors as features for the SVM. This approach resulted in 88.0\% and 59.4\% accuracies on the UCF-101 and HMDB51 datasets, respectively. Due to the impressive performance of this approach, especially for deep learning-based approaches to action recognition, it has seen much interest in recent action recognition research. This novel two-stream approach, as well as the optical flow volume inputs, have been widely studied and adapted to try improve the recognition performance \cite{i3d,potion,tsn1}.

\section{Methodology}
\label{sec:Methodology}

\begin{figure}
\footnotesize
        \centering
\begin{tikzpicture}[auto, node distance=1.5cm,>=latex']

\node (frame31) [block1] {Frame 1};
\node (frame32) [block1, right of=frame31] {Frame 1};
\path (frame31) -- (frame32) node[midway] (tmp2) {};
\node (frame33) [block2, right of=frame32] {Frame 2};
\node (frame34) [block2, right of=frame33] {Frame 2};
\node (frame35) [block3, right of=frame34] {Frame 3};

\node (frame21) [block1, above of=tmp2] {Frame 1};
\node (frame22) [block1, right of=frame21] {Frame 1};
\path (frame21) -- (frame22) node[midway] (tmp1) {};
\node (frame23) [block2, right of=frame22] {Frame 2};
\node (frame24) [block3, right of=frame23] {Frame 3};

\node (frame11) [block1, above of=tmp1] {Frame 1};
\node (frame12) [block2, right of=frame11] {Frame 2};
\node (frame13) [block3, right of=frame12] {Frame 3};

\node [draw=black, fit=(frame11) (frame13)] (bbox1) {};
\node [draw=black, fit=(frame21) (frame24)] (bbox2) {};
\node [draw=black, fit=(frame31) (frame35)] (bbox3) {};

\draw[->, thick] (bbox1) -- (bbox2);
\draw[->, thick] (bbox2) -- (bbox3);
\end{tikzpicture}
\caption{Visualisation of the $t_v < T$ case, for a video of length $3$ frames needing to increase its temporal resolution to $5$ frames. This is to ensure that the logical flow of the action still holds when we need to increase the temporal resolution of a video. This approach is in contrast to the method of simply duplicating the last frame $T - t_v$ times to fill the deficit. We posit that this duplication approach is better for such cases.}
\label{fig:cropfilling}
\end{figure}
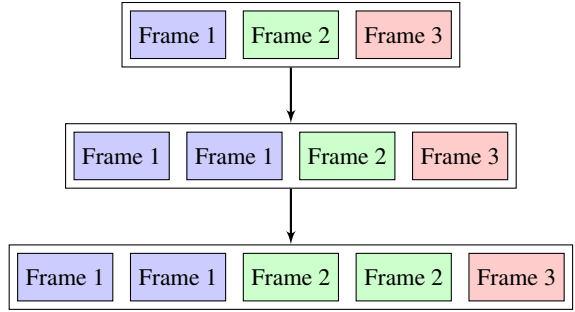

The main intuition behind the proposed method is that actions in video can naturally be decomposed into appearance and motion information, and thus these two components are separately modelled. This is along the same lines as recent research \cite{i3d,tscnn,potion}. These components are modelled using the state-of-the-art 3D CNN architecture known as I3D, employing the RGB network for the spatial modelling, and the flow network for the temporal modelling. One key factor in modelling actions in this way is the spatial and temporal resolutions of the samples used to train the networks. In order to capture as much of the full temporal evolution of the actions as possible, both the spatial and temporal resolutions need to be as high as possible - state-of-the-art techniques using temporal resolutions of $60$ frames or more \cite{i3d,tsn1}. However, \cite{ltc} showed that temporal resolution is somewhat more important than spatial resolution for action recognition. We take cognisance of these findings in our modelling process by using a temporal resolution as large as possible, within computational resource limitations.
Formally, consider a video $V \in \mathcal{G}^{t_v \times r_v \times c_v \times 3}$, where $\mathcal{G} = \{0, 1, \dots, 255\}$ is the gray scale for 8-bit images; $t_v$ is the number of frames in the video; and $r_v \times c_v \times 3$ is the spatial resolution of the frames of the video. We first compute the optical flow video at this original resolution. This is computed by applying dense optical flow for each pair of frames in the video. This flow procedure can be defined as a mapping $h : \mathcal{G}^{r_v \times c_v} \times \mathcal{G}^{r_v \times c_v} \mapsto \mathbb{R}^{r_v \times c_v \times 2}$, which takes in two grayscale frames as inputs, and outputs a $2$-channel `image'. The two channels of this image correspond to the horizontal and vertical components of the flow vector fields, respectively. This results in an optical flow video $F \in \mathbb{R}^{t_f \times r_f \times c_f \times 2}$.
The frames of the RGB video $V$ are normalised by dividing all values by $255$. The temporal resolution is then limited to $T$ frames. This can result in one of $3$ cases. If $t_v < T$, we perform \emph{crop filling}, shown in Figure \ref{fig:cropfilling}. We posit that this duplication process will retain the natural flow and progression of the video or action better than the common alternative in literature of repeatedly appending the last frame to fill the deficit. If $t_v > T$, we sample $T$ equally-spaced frames from $V$. Lastly, if $t_v = T$, we sample the full video. We apply this same temporal sampling to flow video $F$. This results in volumes $V_s \in [0, 1]^{T \times r_v \times c_v \times 3}$ and $F_s \in \mathbb{R}^{T \times r_f \times c_f \times 2}$. Next, all frames of both $V_s$ and $F_s$ are resized such that the smaller side is equal to $S_1$, preserving aspect ratio in the process. This yields volume $V_r$ and $F_r$, respectively.

A rescaling process is then performed on flow video $F_r$, similar to the process performed in \cite{ltc}. The intuition behind the rescaling is that, if the flow for a particular pixel was $l$ at a particular resolution, then the flow at, say, half the resolution, should be $l/2$ (i.e. no longer $l$). Previous works typically do not perform this rescaling, and thus we posit the flow values are not truly representative of the flow at that scale / resolution. This rescaling proved useful for discriminating between temporally-similar actions, such as \emph{walk} and \emph{run}. The rescaling is defined by two factors, $s_x \in \mathbb{R}$ and $s_y \in \mathbb{R}$, representing the scaling factors for $x$ and $y$ flow components, respectively. Formally, given an original spatial resolution of $r_o \times c_o$, and a new resized resolution of $r_n \times c_n$, we have that:
\begin{equation}
s_x =
\begin{cases}
1 - {\left(c_o - c_n\right)}/{c_o} & \text{if $c_n < c_o$} \\
1 + {\left(c_o - c_n\right)}/{c_n} & \text{if $c_n > c_o$} \\
1 & \text{if $c_n = c_o$}
\end{cases}
\end{equation}
Scale factor $s_y$ is computed in a similar manner. The flow is rescaled by multiplying all elements of the $x$ vector fields by $s_x$, and all elements of the $y$ vectors fields by $s_y$.
Finally, we crop $5$ volumes from $V_r$ and $F_r$, where each crop is such that $V_c \in \mathbb{R}^{T \times \frac{7}{8} S_1 \times \frac{7}{8} S_1 \times d_v}$ and $F_c \in \mathbb{R}^{T \times \frac{7}{8} S_1 \times \frac{7}{8} S_1 \times d_f}$, where $V_c$ is a crop from $V_r$, and $F_c$ is a crop from $V_r$. The $5$ crops are the four corners and the centre crop. The coefficient of $7/8$ ensures that the $5$ crops have good spatial coverage of the videos / actions.
Next, we use the RGB and flow I3D networks pre-trained on the ImageNet and Kinetics datasets. We fine-tune the RGB network on the $V_c$ crops, and fine-tune the flow network on the $F_c$ crops. Our baseline approach is taking a consensus vote over the network's predictions for the $10$ crops of a video ($5$ RGB crops, and $5$ flow crops). However, since this consensus voting scheme is naive, and only operates on a crop level, we propose an alternative approach in which the classifier have context about all $5$ crops available to it in order to make its prediction. This context, we posit, will prove beneficial, and aid recognition performance.
The penultimate layer of both I3D networks can be used as generic action features. To this end, we sample $5$ feature vectors for both RGB and flow corresponding to the $5$ crops for both respective streams. Assume these vectors are $D$-dimensional. We then concatenate these $10$ vectors to form one feature vector $\mathbf{v} \in \mathbb{R}^{10D}$. This feature vector is then power-normalised: $\mathbf{v}_{\text{norm}} := \text{sign}(\mathbf{v}) |\mathbf{v}|^{\alpha}$. We use $\alpha = 0.5$ in all experiments. We use power normalisation in place of regular $L_2$-normalisation since it has shown to give better performance \cite{idt}. Since $D$ may be in the order of tens of thousands of dimensions, we first reduce the dimensionality of these vectors using PCA (PCA will be applied by default unless stated otherwise). This reduced-dimension vector is used as the final `action' representation. A linear SVM is then trained on these vectors, and learns to classify actions from this representation. We posit that this will achieve better performance than making predictions directly from the network for three reasons. Firstly, the majority voting scheme from above is simple. Secondly, the layers of the network up until before the final layer can be seen to learn a rich representation of the input that the final layer can classify using a softmax classifier. We posit that an SVM is more effective for classifying from these representations than the softmax classifier. Lastly, the SVM will have the information about all RGB and flow crops from the $\mathbf{v}$ representation, whereas the network makes predictions on a crop-by-crop basis, and thus does not have the information from other crops to aid its prediction.
\subsection{Hyperparameter Optimisation}
For the linear SVMs, the only parameter we optimise for is the penalty parameter $C$. We use a grid search approach, and choose the $C$ that maximises five-fold cross-validation accuracy. The parameters for the I3D networks, such as number of layers and number of filters / neurons per layer, are left as default as per the recommendation of the authors in \cite{i3d}. The weights for the network are set to that setting which minimises the cross-entropy loss on a validation dataset.

\section{Experimental Results}
\label{sec:Experimental Results}

\subsection{Action Recognition Datasets}
In this research, we study and apply our proposed technique to the KTH and HMDB51 datasets. This provides a good framework to test our method on a simple dataset, KTH (as a sort of regression test), and a more difficult, complex dataset, HMDB51.

The KTH dataset consists of fairly-static backgrounds, with actions performed 25 different actors. It consists of 599 videos of 25 people repeatedly performing one of the six actions. Altogether, of which there are 2391 subsequences of actions being performed. There are four different scenarios spread across the videos, namely \emph{s1}: outdoors; \emph{s2}: outdoors with scale variation; \emph{s3}: outdoors with different clothes; and \emph{s4}: indoors. Further, the dataset consists of the following six action classes: Walking; Jogging; Running; Boxing; Handclapping; and Handwaving.
HMDB51 \cite{hmdb51} is a large human action recognition dataset, often used to thoroughly test a particular approach to action recognition, as it is widely considered one of the more difficult benchmark datasets in this domain. It consists of videos from a variety of sources such as films, public databases, YouTube, and Google videos. There are 6676 clips divided into 51 different categories/classes. Each class contains a minimum of 101 videos. Rather than enumerate all 51 action classes, below is a description of the 5 general classes that the 51 categories are grouped into \cite{hmdb51}: General facial actions; Facial actions with body manipulations; General body movements; Body movements with object interactions; and Body movements for human interaction.

UCF101 \cite{ucf101} is a large-scale human action recognition dataset, consisting of 13320 realistic videos collected from YouTube. There are 101 action classes in the dataset, which can be broadly grouped into the following 5 categories: Human-Object Interaction; Body-Motion Only; Human-Human Interaction; Playing Musical Instruments; and Sports.

The videos contain large variations in illumination, scale, viewpoint, appearance, and pose, and the backgrounds are often cluttered. There exists some consistency between videos (such as a common background or viewpoint), even between those which are in separate classes.
\subsection{Performance Evaluation}
For both datasets, we will use the average accuracy over the classes as the main metric for evaluating performance. This is the standard evaluation metric from previous literature. For the KTH dataset, the data is split according to the standard split of 8 people for training, 8 for validation, and 9 for testing. We then average the results over a pre-defined number of trials of this train/test split framework. Performance is then measured by average accuracy over these splits. For the HMDB51 dataset, there are three pre-defined train/test splits that come with the dataset, as defined by the introducing authors. Performance is then measured by average accuracy over these three splits. We also provide performance measures in the form of classification performance tables (showing class-by-class precision, recall, and F1-score), and confusion matrices.

\subsection{Experimental Setup}
Technologies used to run the experiments are Python with various machine learning, computer vision, and deep learning libraries. Experiments were run on a 32-core machine with 128GB of RAM and a RTX 2080Ti GPU, as well as a separate 8-core machine, with 32GB of RAM, and a GTX 1080 GPU.
\subsection{Results on KTH Dataset}
To test on the KTH dataset we set parameter $S_1 = 128$ and parameter $T = 40$. Such a setting, we posit, should be able to adequately model the comparatively-simple 2391 snippets of the dataset. We split the dataset in the recommended manner of 8 people for training, 8 for validation, and 9 for testing. We fine-tune the I3D networks for a certain number of epochs, and use the model that minimises the cross-entropy loss on the validation set.
Due to the KTH dataset's relative simplicity, we would expect performance to be fairly high. We can see a comparison of our proposed approaches against previous methods in Table \ref{tbl:kthacc}. It is clear that our method is competitive with state-of-the-art approaches, achieving a highest accuracy of 96.6\%, which is second only to the state-of-the-art approach by \cite{corso1} with 98.2\%. Even with the baseline majority voting approach, performance is still competitive. We can most likely attribute this performance to the KTH's simplicity. The backgrounds are mostly static, and homogeneous. There is also not much clutter and occlusion to contend with. The I3D networks are, therefore, able to model the actions from such videos well. The ActionBank method proposed by \cite{corso1} performs better most likely due to the amount of domain knowledge incorporated into the hand-crafted features, however, it should be noted that this method does not generalise well to complex settings / datasets such as HMDB51. Our method is, instead, simply tasked with learning the actions directly from RGB pixel values and optical flow vector fields, and generalises better than ActionBank to such settings. The baseline approach is less powerful simply because it makes predictions on a crop-by-crop basis and performs a majority vote. There are no steps taken to use information from the other crops (e.g. in the form of pooling or the SVM approach) to make the predictions. However, it is still relatively high, since much of the context about an action get be determined from a single video crop (as a result of the simple background, minimal noise, and limited clutter / occlusion).

Our approach performs better than the other methods because our method is structured in such a way that it makes no stringent assumptions about the actions. Instead, we feed raw RGB and optical flow inputs and task the networks with learning everything. The majority of the previous approaches hand-craft features that may work well on some simpler datasets, but struggle to generalise to more complex settings. Our method has no such limitation, and as such has a higher accuracy. Some of the previous works do compete with our method, such as \cite{densetraj} and \cite{grauman1}. The dense trajectory representation \cite{densetraj} is strong since it extracts both zero-order (HOF) and first-order (MBH) motion information, as well as trajectory and appearance (HOG) information. It then encodes and quantises these descriptors using the state-of-the-art Fisher vector representation. This means the representation is powerful enough to perform well on the KTH dataset since it takes many different cues into account to represent an action (and similarly for the hierarchical representation employed in \cite{grauman1}, where the features are able to encode enough pertinent information about the actions to perform well).
\begin{table}
\centering
\small
\caption{Results on the KTH dataset.}
\label{tbl:kthacc}
\begin{tabular}{|l|l|}
\hline
\textbf{Method} & \textbf{Accuracy} \\ \hline \hline
N-Jets + BoW \cite{laptev2}       & \multicolumn{1}{c|}{71.7}              \\ \hline
STIP + HOG/HOF \cite{laptev5}          & \multicolumn{1}{c|}{91.8}              \\ \hline
Dense Sampling + HOG/HOF \cite{laptev3}            & \multicolumn{1}{c|}{92.1}              \\ \hline
Action Bank \cite{corso1}           & \multicolumn{1}{c|}{98.2}              \\ \hline
Action MACH \cite{shah}            & \multicolumn{1}{c|}{88.66}             \\ \hline
Hough Forest \cite{yao1}    & \multicolumn{1}{c|}{92.0}              \\ \hline
3D HOG \cite{klaser2}         & \multicolumn{1}{c|}{91.4}              \\ \hline
Space-Time Hierarchy \cite{grauman1}           & \multicolumn{1}{c|}{95.53}             \\ \hline
ISA \cite{ng1}             & \multicolumn{1}{c|}{93.9}              \\ \hline
Dense Trajectories \cite{densetraj}              & \multicolumn{1}{c|}{94.53}             \\ \hline
\textbf{Ours - baseline}              &  \multicolumn{1}{c|}{\textbf{94.5}}            \\ \hline
\textbf{Ours - SVM}              &  \multicolumn{1}{c|}{\textbf{96.6}}          \\ \hline
\end{tabular}
\end{table}

We ran an additional experiment to investigate the effect of PCA on performance. Surprisingly, reducing the dimension from the original 20400 dimensions to approximately 1500 dimensions resulted in a marginally higher accuracy (i.e. the system's performance is slightly worse \emph{without} PCA). This is likely due to the curse of dimensionality, although the difference is so negligible, that we are unable to draw any real conclusions. These very large dimensions also discount the use of non-linear kernels in the SVM, as they typically do not perform well in such contexts. Moreover, linear SVMs have been shown to perform well in a high-dimensional action recognition context \cite{densetraj,idt}.

As part of our method, we perform what we term as `crop filling', which is appending frames to a crop of a video if the number of frames is less than the desired number for the input into the networks. This is typically done by repeatedly appending the last frame the required number of times until the deficit is filled. We instead propose a scheme as described in the previous chapter, and can be seen in Figure \ref{fig:cropfilling}. Additionally, we also perform optical flow scaling as opposed to leaving the flow values untouched as often done in previous research. We investigate leaving out these two steps - that is, appending the last frame and performing no flow scaling. This resulted in a baseline accuracy of 91.5\%, which is noticeably lower than the baseline accuracy compared with when the two proposed methods are employed - 95.0\%. This suggests that performing these two steps aids performance. Interestingly, the performance of the SVM method drops only very slightly to 95.7\%. Since it is difficult to tell how much these two techniques help on the KTH dataset, we will see that they are indeed beneficial in more complex settings, such as on the HMDB51 dataset. We can see aforementioned results in Table \ref{tbl:kthscalefill}.

\begin{table}
\centering
\small
\caption{Results of the proposed method with and without the flow scaling and crop filling procedures on the KTH dataset.}
\label{tbl:kthscalefill}
\begin{tabular}{|l|l|l|}
\hline
                                     & \textbf{Baseline} & \textbf{SVM}  \\ \hline \hline
No Flow Scaling + Crop Filling       & 91.5              & 95.7          \\ \hline
\textbf{Flow Scaling + Crop Filling} & \textbf{95.0}     & \textbf{95.8} \\ \hline
\end{tabular}
\end{table}

\subsection{Results on HMDB51 Dataset}
To test on the HMDB51 dataset, we set parameter $S_1 = 256$ and $T = 40$. This setting is partially inspired by the findings by \cite{ltc}, in which it is shown that temporal resolution is somewhat more important in action recognition than spatial resolution. Thus, we attempt to make the temporal resolution as long as possible, within the constraints of computational resource (and time) limitations (40 being the limit in the circumstances). A large temporal resolution is needed for complex, varietal datasets such as HMDB51 since the videos are fairly long, and, more importantly, the majority of the full temporal evolution of the action needs to be modelled to ensure enough of the variation in the data is being captured. The latter is because the dataset has some actions that are \emph{very} temporally (and spatially) similar - \emph{talk} vs \emph{laugh}, \emph{draw sword} vs \emph{sword exercise}, and \emph{kick} vs \emph{kick ball}. This is in tandem with the other existing difficulties related to the dataset, such as very complicated, non-static backgrounds, clutter, and occlusion. Performance for this dataset is represented as accuracy. We fine-tune the I3D networks for a certain number of epochs, and use the model that minimises the cross-entropy loss on the validation set, where the validation set is given by 20\% of the training data.

Results on the HMDB51 dataset can be seen in Table \ref{tbl:hmdb51acc}. It is clear that the method is not competitive with state-of-the-art approaches \cite{i3d} and \cite{potion}. We attribute this to computational resource limitations, as we were not able to train for temporal (and spatial) resolutions as large as those approaches used (i.e. $>= 64$ frames). This makes a large difference since this is a 60\% increase in temporal resolution that the networks can learn from, which is vital for complex datasets such as HMDB51 (and less so for a dataset such as KTH). The authors from these approaches had more than 60 GPUs to train their networks, whereas we had only 1 available. Time available for the research was also a prohibitive factor. Further, these state-of-the-art approaches employ the more accurate TVL-1 optical flow algorithm to compute the optical flow videos, whereas we instead opt for Farneback's optical flow algorithm. We do this since the TVL-1 algorithm is very computationally intensive and prohibits such systems' real-time capabilities (even though the resulting flow estimates are more accurate). However, it should be noted that the goal of this work is not to necessarily to achieve state-of-the-art performance on these datasets, but rather to thoroughly investigate using RGB (spatial) and flow (temporal) information to model actions in video, and introduce potential improvements to the baseline, such as optical flow scaling and using another class of model on the extracted features of the networks (in our case, a linear SVM). These improvements may prove useful in competing with state-of-the-art when these networks can be trained on larger batches of crops with higher temporal and spatial resolutions. This, however, is left as future work.
\begin{table}[!t]
\centering
\small
\caption{Results on the HMDB51 dataset.}
\label{tbl:hmdb51acc}
\begin{tabular}{|l|l|}
\hline
\textbf{Method}       & \textbf{Accuracy} \\ \hline \hline
Action Bank \cite{corso1}                 & \multicolumn{1}{c|}{26.9}              \\ \hline
iDT + FV \cite{idt}                   & \multicolumn{1}{c|}{57.2}              \\ \hline
MPEG Flow \cite{kantorov1}             & \multicolumn{1}{c|}{46.7}              \\ \hline
iDT + Stack Fisher vectors \cite{sfv1}                   & \multicolumn{1}{c|}{66.7}              \\ \hline
Two-Stream CNN \cite{tscnn}                    & \multicolumn{1}{c|}{59.4}              \\ \hline
Two-Stream CNN + Fusion \cite{feichtenhofer2016convolutional}                    & \multicolumn{1}{c|}{65.4}              \\ \hline
Long-Term Temporal Convolution \cite{ltc}                    & \multicolumn{1}{c|}{64.8}              \\ \hline
I3D \cite{i3d}                   & \multicolumn{1}{c|}{80.9}              \\ \hline
PoTion + I3D \cite{potion}                & \multicolumn{1}{c|}{80.9}              \\ \hline
\textbf{Ours - baseline} & \multicolumn{1}{c|}{\textbf{50.7}}     \\ \hline
\textbf{Ours - SVM} & \multicolumn{1}{c|}{\textbf{62.8}}     \\ \hline
\end{tabular}
\end{table}

We ran an additional experiment to investigate the effect of not applying PCA to the features before normalising and feeding them into the SVM. The results without performing PCA are negligible. The dimension is reduced from 20400 to 3500. This suggests that there is a low-dimensional basis governing these local CNN features, as we can reduce the number of dimensions by over 80\% and see no difference in performance. Also, we can conclude that, for the HMDB51 dataset, applying PCA is a better idea than not applying it, as we are left with fewer dimensions to deal with, and there will be a resultant reduction in training and testing time (with no effect on performance) for the linear SVM.

Similarly to the KTH dataset, we perform an investigation of not performing the `crop filling' and flow scaling procedures. This resulted in a baseline accuracy of 48.4\%, which is lower than the baseline accuracy compared with when the two proposed methods are employed - 50.6\%. The SVM-based approach also drops from 64.1\% to 59.9\%. This reinforces the suggestion that performing these two steps aids performance, and the benefits of doing so are more clear on the HMDB51 dataset than on the KTH dataset. We posit that the scaling procedure results in flow values that are more representative of what the flow values should be at the resolution of the input. The `filling' procedure, we posit, represents a more natural progression of the video, and by extension, the action, than simply repeatedly appending the last frame would. We can see a tabulation of these aforementioned results in Table \ref{tbl:hmdb51scalefill}.
\begin{table}[!t]
\centering
\small
\caption{Results of the proposed method with and without the flow scaling and crop filling procedures on the HMDB51 dataset (split 1).}
\label{tbl:hmdb51scalefill}
\begin{tabular}{|l|l|l|}
\hline
                                     & \textbf{Baseline} & \textbf{SVM}  \\ \hline \hline
No Flow Scaling + Crop Filling       & 48.4              & 59.9          \\ \hline
\textbf{Flow Scaling + Crop Filling} & \textbf{50.6}     & \textbf{64.1} \\ \hline
\end{tabular}
\end{table}

\subsection{Results on the UCF101 Dataset}
For testing on the UCF101 dataset, we set parameter $S_1 = 176$ and $T = 30$. This setting is the maximum possible for our time and computational resource limitations. It should be noted, however, that previous works used significantly larger spatial and temporal resolutions for this dataset. Large temporal and spatial resolutions are very important for relatively complex datasets such as UCF101. It should be noted that our quoted accuracy figures for the UCF101 dataset are for split 1 only.

It is clear in Table \ref{tbl:ucf101acc} that the proposed SVM on the full video context, as opposed to the baseline softmax classifier on a portion of the context, significantly improves performance. This is the same trend as seen in the other two studied benchmark datasets. Overall performance is lacking compared to the state-of-the-art approaches for the same reasons discussed for the HMDB51 dataset. Computational resources limitations prevented us from obtaining maximum performance from the I3D network, which therefore inherently limits the representational power of the extracted features for RGB and flow. As shown in \cite{ltc}, greater spatial and temporal resolutions significantly improve performance of action recognition systems. However, for our relatively small spatial and temporal resolutions, we empirically show that the SVM approach does indeed prove beneficial.
\begin{table}[!t]
\centering
\small
\caption{Results on the UCF101 dataset.}
\label{tbl:ucf101acc}
\begin{tabular}{|l|l|}
\hline
C3D \cite{c3d}                    & \multicolumn{1}{c|}{90.4}              \\ \hline
Two-Stream CNN \cite{tscnn}                    & \multicolumn{1}{c|}{88.0}              \\ \hline
Two-Stream CNN + Fusion \cite{feichtenhofer2016convolutional}                    & \multicolumn{1}{c|}{93.5}              \\ \hline
Long-Term Temporal Convolution \cite{ltc}                    & \multicolumn{1}{c|}{92.7}              \\ \hline
I3D \cite{i3d}                   & \multicolumn{1}{c|}{98.0}              \\ \hline
PoTion + I3D \cite{potion}                & \multicolumn{1}{c|}{98.2}              \\ \hline
\textbf{Ours - baseline} & \multicolumn{1}{c|}{\textbf{80.5}}     \\ \hline
\textbf{Ours - SVM} & \multicolumn{1}{c|}{\textbf{86.5}}     \\ \hline
\end{tabular}
\end{table}

In Table \ref{tbl:ucf101scalefill}, we see an interesting result that does not follow the trend of the HMDB51 and KTH datasets. That is, for this dataset - UCF101 - the flow scaling and crop filling does not really aid performance. Performance stays pretty much the same for both the baseline and SVM approaches. A likely cause for this is the fact that, firstly, the crop filling case of $t_v < T$ is a rare in a dataset with relatively long videos such as in UCF101. Additionally, flow scaling typically helps most when distinguishing between actions that are very temporally similar. UCF101 has fewer of these in comparison to HMDB51 and KTH. Most of the actions in the dataset are temporally dissimilar.

\begin{table}[!t]
\centering
\small
\caption{Results of the proposed method with and without the flow scaling and crop filling procedures on the UCF101 dataset (split 1).}
\label{tbl:ucf101scalefill}
\begin{tabular}{|l|l|l|}
\hline
          & \textbf{Baseline} & \textbf{SVM}  \\ \hline \hline
\textbf{No Flow Scaling + Crop Filling}       & \textbf{80.5}              & \textbf{86.4}          \\ \hline
Flow Scaling + Crop Filling & 79.8     & 86.3 \\ \hline
\end{tabular}
\end{table}

\section{Conclusion and Future Work}
\label{sec:Conclusion and Future Work}

Given the limited temporal resolution we were able to employ, the method performs admirably. It, unsurprisingly, competes with state-of-the-art on the KTH dataset, achieving a highest average accuracy of $96.6$\%. The simple backgrounds, and lack of clutter and occlusion in the videos mean performance on the dataset is high. We outperform all previous methods exception that of \cite{corso1}. However, we significantly outperform \cite{corso1} on the realistic, complex dataset HMDB51, suggesting our method is able to generalise to these settings more effectively. Due to the limited temporal (and spatial) resolution, we are not able to compete with state-of-the-art on the HMDB51 and UCF101 datasets, achieving a highest average accuracy of $62.8$\%, and $86.4$\%, respectively. However, we still compete with other deep learning-based methods, and outperform many engineered features approaches. More importantly, we are able to effectively demonstrate our two main goals of the research. Firstly, using a SVM trained on context from all crops of the video is significantly more effective than taking a majority vote of crop-level network predictions. The benefit of the SVM is more apparent on the HMDB51 and UCF101 datasets since, for the KTH dataset, much of the action's context can be determined from a single crop (whereas this does not apply for the more complex HMDB51 and UCF101 datasets). Secondly, performing the two simple pre-processing steps of crop filling and optical flow scaling results in higher recognition performance for the KTH and HMDB51 datasets, since these datasets contain more temporally similar actions than UCF101.

The most important extension to this work is to increase the temporal resolution (and spatial resolution) to see how this affects performance, and at one point does performance plateau - when the increase in resolution no longer aids performance. Furthermore, different, more accurate, optical flow algorithms could be used in place of Farneback's methods (such as TVL-1 \cite{tvl1} or Brox \cite{brox}). However, these methods are generally much slower to compute, and thus to make them feasible for a real-time human action recognition system one would likely need to use their GPU implementations. Another possible extension is to train the networks on more crops of the videos (by sampling more crops, or not limiting the temporal resolution and then sampling from the full video).

\bibliographystyle{unsrt}  
\bibliography{references}  






\end{document}